\documentclass[conference]{IEEEtran}
\IEEEoverridecommandlockouts

\usepackage{graphicx}
\usepackage{tikz}
\usetikzlibrary{positioning}
\usepackage{amsmath, amssymb}
\usepackage{url}
\usepackage{hyperref}
\usepackage{booktabs}
\usepackage{caption}
\usepackage{float}
\usepackage{cite}
\usepackage{subcaption} 

\begin{document}

\title{Oitijjo-3D: Generative AI Framework for Rapid 3D Heritage Reconstruction from Street View Imagery}

\author{
\IEEEauthorblockN{
\begin{tabular}{@{}c@{\hspace{2em}}c@{\hspace{2em}}c@{}}
\textbf{Momen Khandoker Ope} & \textbf{Akif Islam} & \textbf{Mohd Ruhul Ameen} \\
\textit{University of Rajshahi, Bangladesh} & \textit{University of Rajshahi, Bangladesh} & \textit{Marshall University, USA} \\
\texttt{khandokermomen919@ru.ac.bd} & \texttt{iamakifislam@gmail.com} & \texttt{ameen@marshall.edu} \\
[0.8em]
\textbf{Abu Saleh Musa Miah} & \textbf{Md Rashedul Islam} & \textbf{Jungpil Shin} \\
\textit{University of Aizu, Japan} & \textit{University of Asia Pacific, Bangladesh} & \textit{University of Aizu, Japan} \\
\texttt{musa@u-aizu.ac.jp} & \texttt{rashed.cse@gmail.com} & \texttt{jpshin@u-aizu.ac.jp}
\end{tabular}}
}

\maketitle

\begin{abstract}
Cultural heritage restoration in Bangladesh faces a dual challenge of limited resources and scarce technical expertise. Traditional 3D digitization methods—such as photogrammetry or LiDAR scanning—require expensive hardware, expert operators, and extensive on-site access, which are often infeasible in developing contexts. As a result, many of Bangladesh’s architectural treasures, from the Paharpur Buddhist Monastery to Ahsan Manzil, remain vulnerable to decay and inaccessible in digital form. This paper introduces \textbf{Oitijjo-3D}, a cost-free generative AI framework that democratizes 3D cultural preservation. By using publicly available \textit{Google Street View} imagery, Oitijjo-3D reconstructs faithful 3D models of heritage structures using a two-stage pipeline - multimodal visual reasoning with Gemini 2.5 Flash Image for structure–texture synthesis, and neural image-to-3D generation through Hexagen for geometry recovery. The system produces photorealistic, metrically coherent reconstructions in seconds—achieving compared to conventional structure,from Motion pipelines—without requiring any specialized hardware or expert supervision. Experiments on landmarks such as Ahsan Manzil, Choto Sona Mosque, and Paharpur demonstrate that Oitijjo-3D can preserve visual and structural fidelity while drastically lowering economic and technical barriers. By turning open imagery into digital heritage, this work reframes preservation as a community-driven, AI-assisted act of cultural continuity for resource-limited nations.
\end{abstract}


\section{Introduction}

\begin{figure*}[!t]
\centering
\includegraphics[width=0.32\textwidth, height=0.32\textwidth]{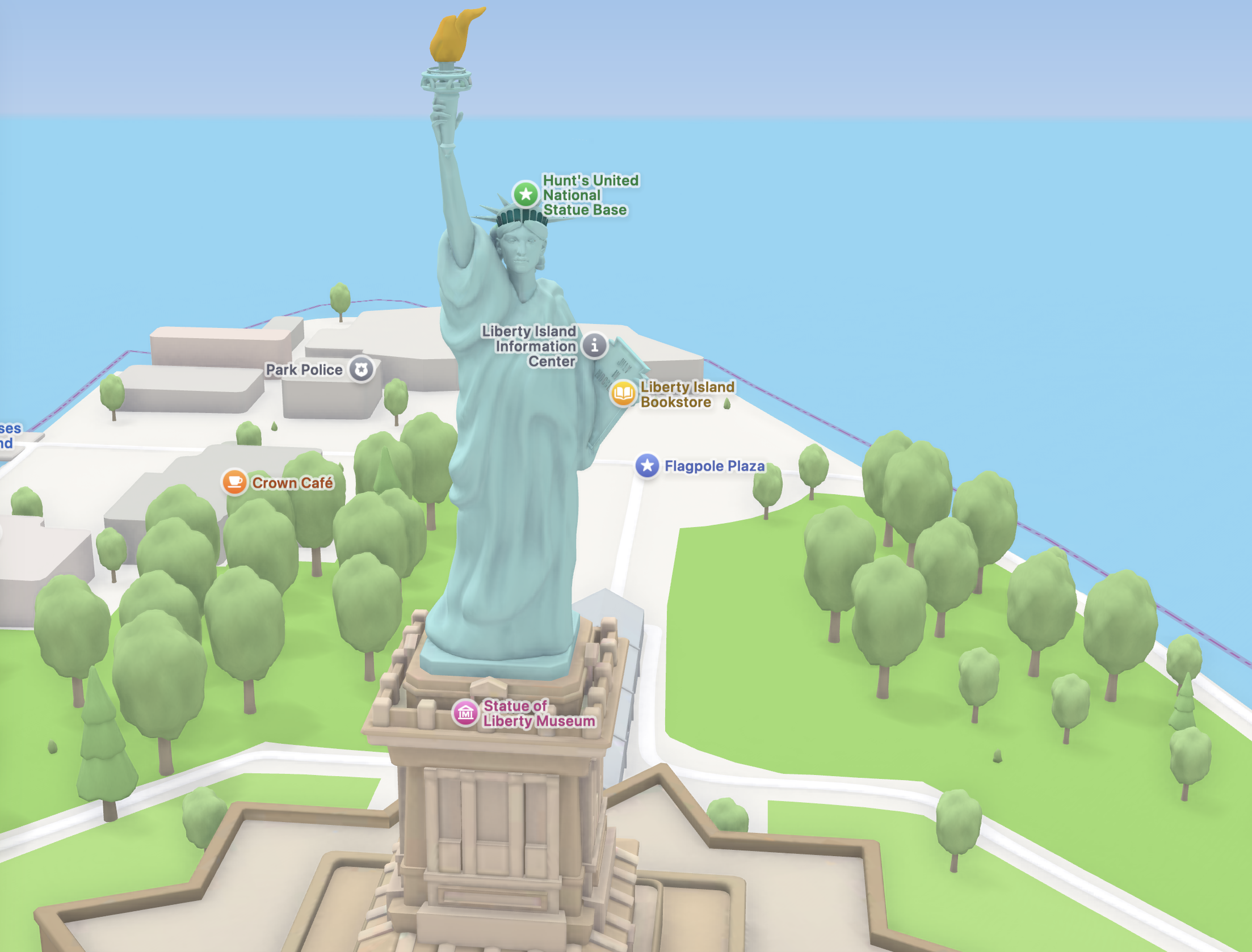} \hfill
\includegraphics[width=0.32\textwidth, height=0.32\textwidth]{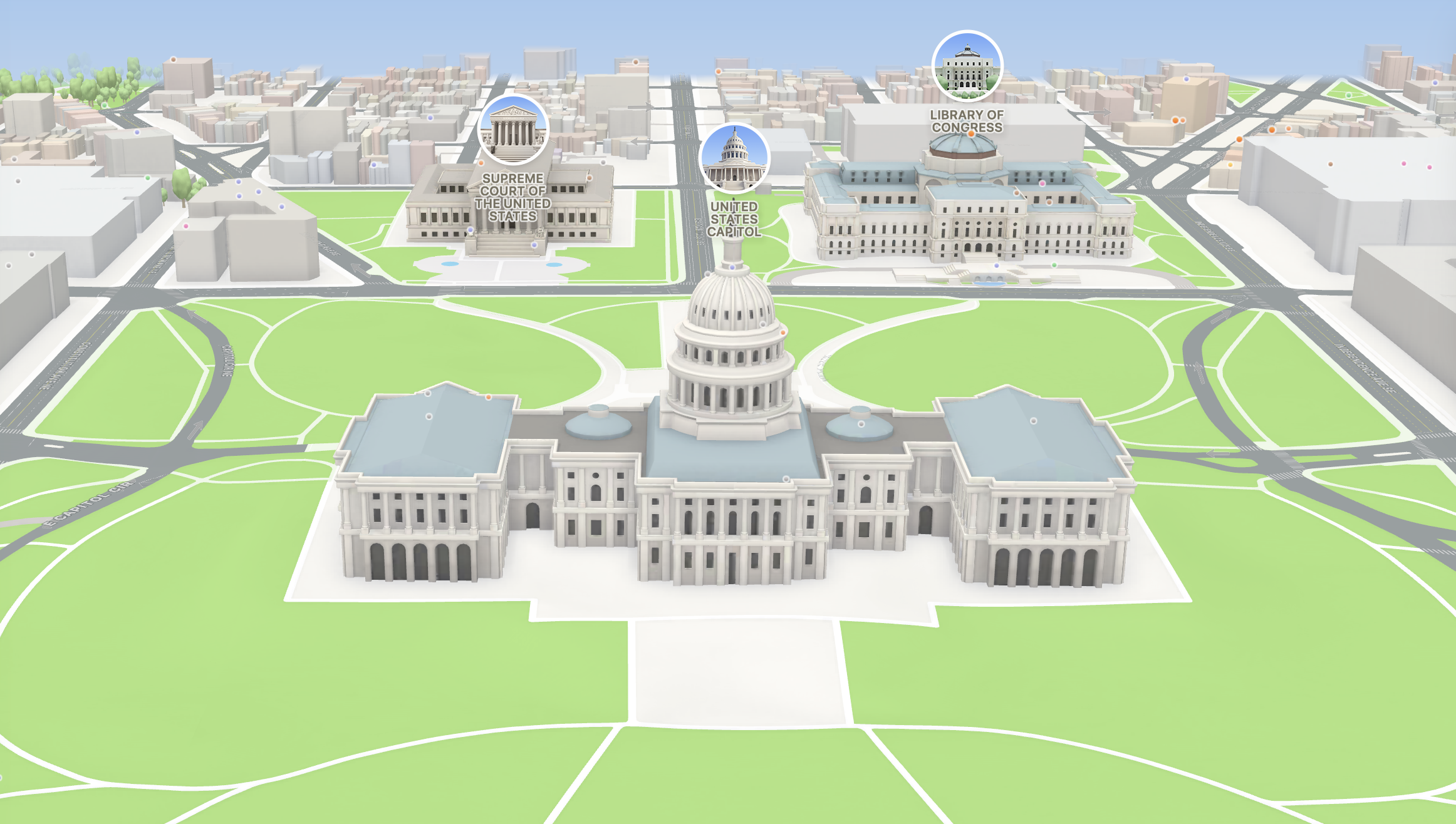} \hfill
\includegraphics[width=0.32\textwidth, height=0.32\textwidth]{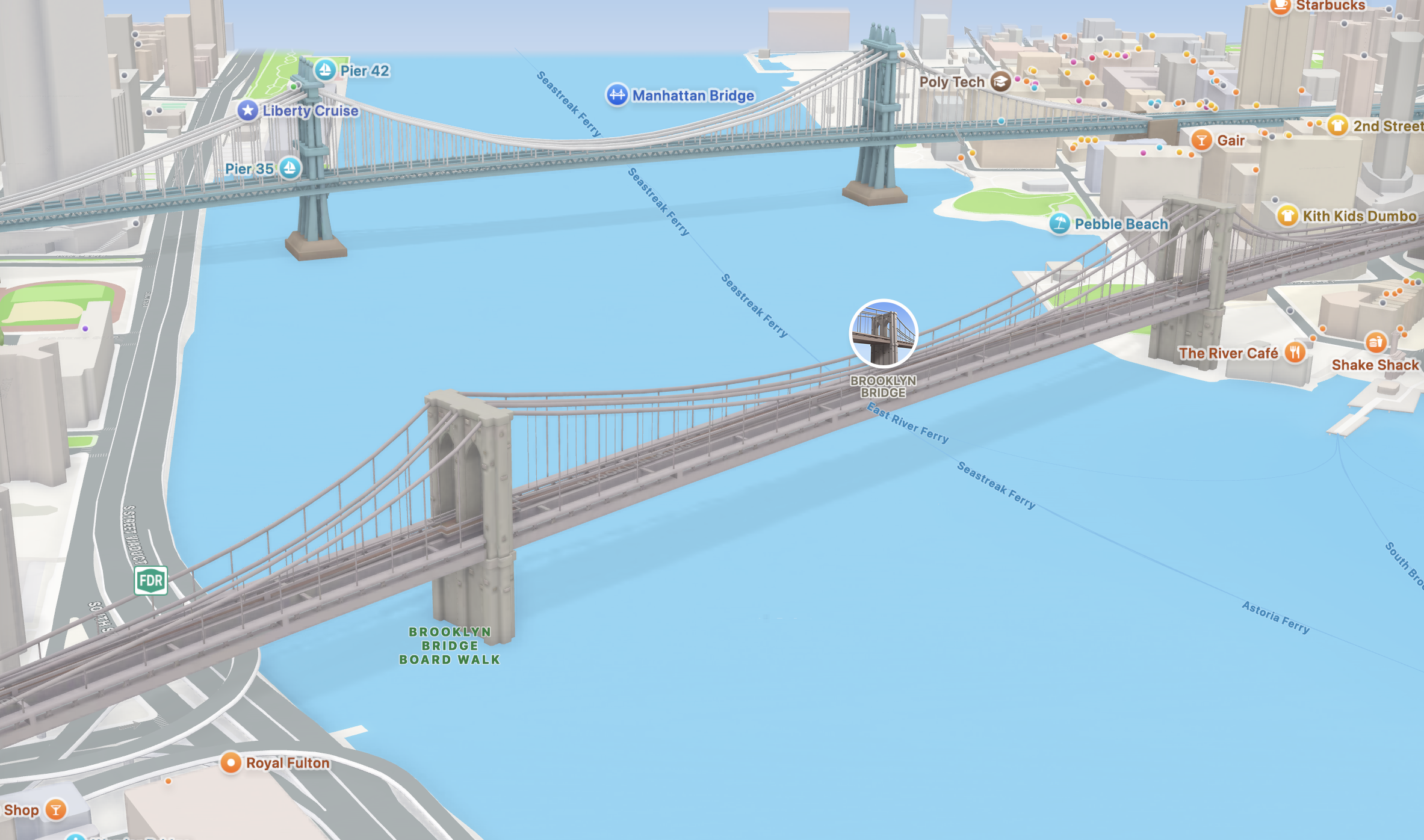}
\caption{Apple Maps 3D feature showing (left) the Statue of Liberty, (center) the United States Capitol, and (right) the Brooklyn Bridge. Such high-fidelity 3D representations are available in the United States but remain absent in underdeveloped countries like Bangladesh.}
\label{fig:apple_3d_examples}
\end{figure*}

Across the streets of Bangladesh, history lives in stone, in the domes of ancient mosques, the arches of colonial palaces, and the quiet ruins of monasteries that once thrived with life. These landmarks are more than architecture; they are the nation’s collective memory. Yet while the world moves into immersive digital experiences, many of these treasures still exist online only as flat, two-dimensional photographs. On Apple Map, one can freely explore the 3D models of the Eiffel Tower or the Statue of Liberty — but not the Kantajew Temple, Paharpur Buddhist Monastery, or Ahsan Manzil Palace (see Figure~\ref{fig:apple_3d_examples}). The contrast is not of cultural significance, but of access and opportunity.

Creating digital 3D models of real-world structures has traditionally required expensive equipment such as LiDAR scanners or dense photogrammetry setups. These methods demand thousands of photographs, technical expertise, and weeks of post-processing — costs that are impossible to bear for most local institutions in underdeveloped countries \cite{calantropio2018lowcost, dhonjua2017feasibility, wikipedia3dscanning}. As a result, an entire generation of South Asian architecture remains digitally invisible, despite being visually documented every day through smartphones, Street View captures, and social media posts.

This paradox — abundance of images but absence of 3D representation — inspired the vision behind \textbf{Oitijjo-3D}. If people already take countless 2D photographs of heritage sites, could those same images be transformed into interactive 3D models using artificial intelligence? Could modern generative models bridge the technological divide that keeps cultural memory out of the digital world?

Recent progress in multimodal AI and diffusion-based systems has made this vision achievable. Models such as Gemini 2.5 Flash Image and Hexagen can now infer geometry, depth, and material texture from a single image, generating lifelike 3D meshes within seconds \cite{liu2023zero1to3, liu2023anysingleimageto3d}. Yet, these capabilities have never been directed toward the preservation of heritage in low-resource contexts — until now.

\textbf{Oitijjo-3D} is an app which introduces a generative AI framework that reconstructs three-dimensional models of historical monuments directly from publicly available \textit{Google Street View} imagery. The system first uses Gemini 2.5 Flash Image model (commonly known as Nano Banana) to generate isometric architectural views from simple image prompts, and then employs Hexagen’s neural depth modeling to transform these into textured 3D meshes.

The novelty of Oitijjo-3D lies not only in its technology but in its philosophy — that heritage preservation should not be a luxury reserved for nations with vast resources. By showing that faithful 3D reconstructions can be generated from the imagery we already have, this work opens a new path for cultural conservation: one that is inclusive, affordable, and powered by the collective data of everyday people.

\section{Related Work}

For decades, researchers have relied on classical 3D reconstruction pipelines such as Structure-from-Motion (SfM) and Multi-View Stereo (MVS) \cite{schonberger2016sfm, furukawa2015multi}. These methods can produce highly accurate geometric models but only under ideal conditions—multiple overlapping images, controlled lighting, and precise calibration. In practice, they are expensive, slow, and often fail when dealing with noisy outdoor scenes or historical structures with uneven textures \cite{remondino2011image}. For most heritage institutions in developing countries, the requirement of high-end cameras, expert supervision, and long computation time simply makes these systems unattainable. The emergence of Neural Radiance Fields (NeRF) \cite{mildenhall2021nerf} marked a new era in 3D scene representation, producing stunning visual fidelity through neural rendering. Yet, NeRF and its variants demand extensive GPU resources and hours of optimization for even a single site. Such dependence on dense multi-view data and powerful infrastructure makes them unsuitable for regions like South Asia, where heritage digitization must be low-cost and easily deployable.

Generative diffusion models have further revolutionized visual synthesis, with systems like GLIDE \cite{nichol2021glide}, Imagen \cite{saharia2022photorealistic}, and Stable Diffusion \cite{rombach2022ldm} creating photorealistic 2D imagery from text prompts. However, their creativity stops at the image plane—these models excel in visual storytelling but lack true 3D spatial understanding. For cultural preservation, where proportional accuracy and structural realism are crucial, such models fall short Single-image-to-3D approaches such as DreamFusion \cite{poole2023dreamfusion}, Zero-1-to-3 \cite{zhou20233d}, and MVDream \cite{shi2023mvdream} have pushed the boundary closer to practical use, generating meshes directly from limited inputs. Still, these methods were built for synthetic objects, gaming assets, or research benchmarks—not for complex architectural forms that define heritage sites. They rarely consider the symmetry, scale, or material context that make monuments authentic.

In heritage research, most previous efforts remain grounded in photogrammetry or LiDAR-based documentation \cite{grussenmeyer2002documentation, hassani2016reviewing}. While technically precise, these methods are prohibitively expensive and require field expertise often unavailable in developing nations. More importantly, despite Bangladesh’s rich architectural history, no prior work has explored a generative, AI-based pipeline for reconstructing its historical landmarks in 3D using publicly available imagery. This absence defines the core research gap that \textbf{Oitijjo-3D} seeks to address. Our novelty lies in harnessing freely available resources—particularly \textit{Google Street View} imagery and community-contributed photographs—to enable 3D reconstruction without costly equipment, manual design effort, or high computational demand. By integrating modern diffusion-based 2D-to-3D generation within this framework, our work demonstrates that realistic and visually faithful reconstructions of Bangladeshi heritage can be achieved quickly, affordably, and accessibly. In doing so, \textbf{Oitijjo-3D} turns cultural preservation from a privilege of well-funded institutions into a practical, open, and inclusive process for all.

\begin{figure*}[!t]
\centering

\begin{subfigure}{0.78\textwidth}
    \centering
    \includegraphics[width=0.32\linewidth, keepaspectratio]{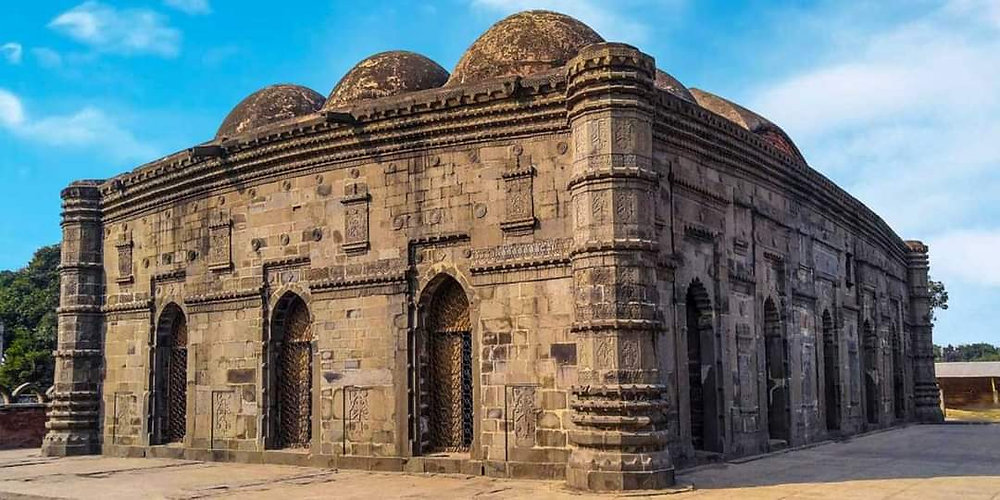}\hfill
    \includegraphics[width=0.32\linewidth, keepaspectratio]{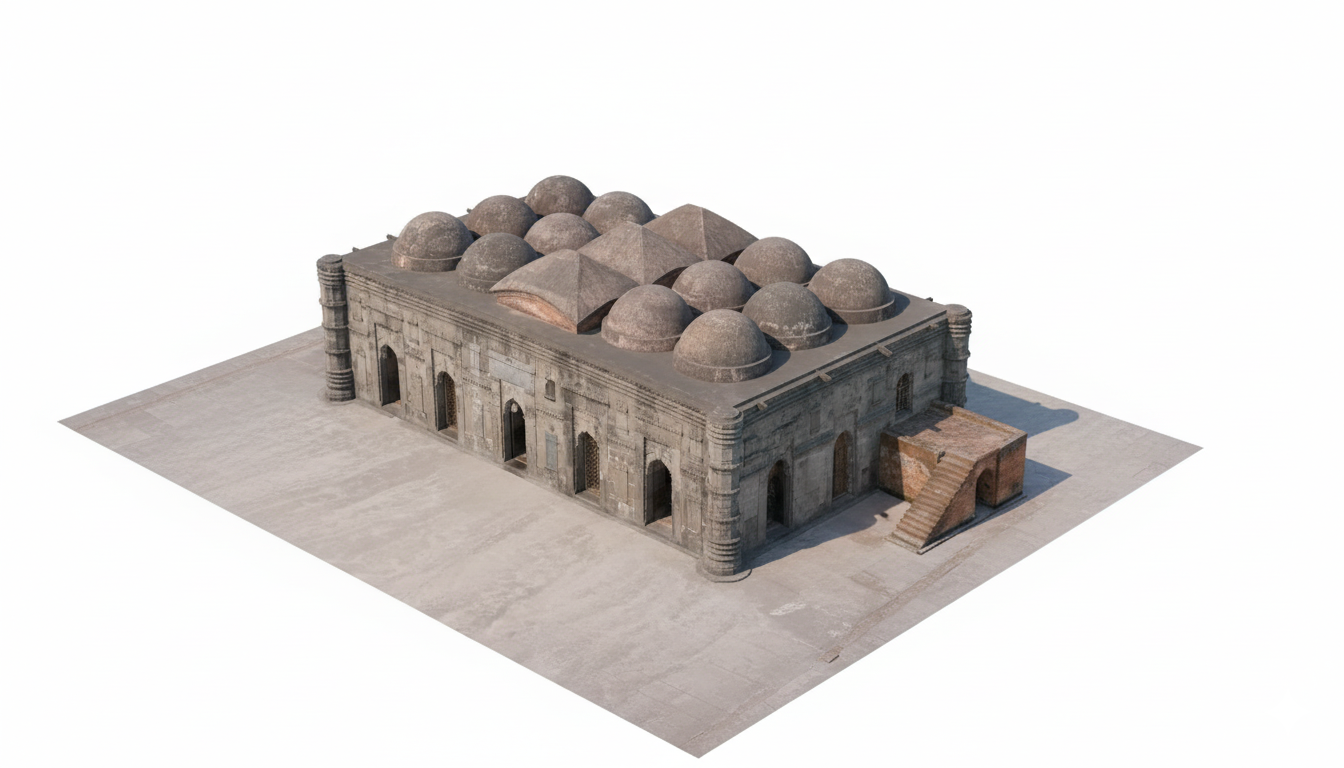}\hfill
    \includegraphics[width=0.32\linewidth, keepaspectratio]{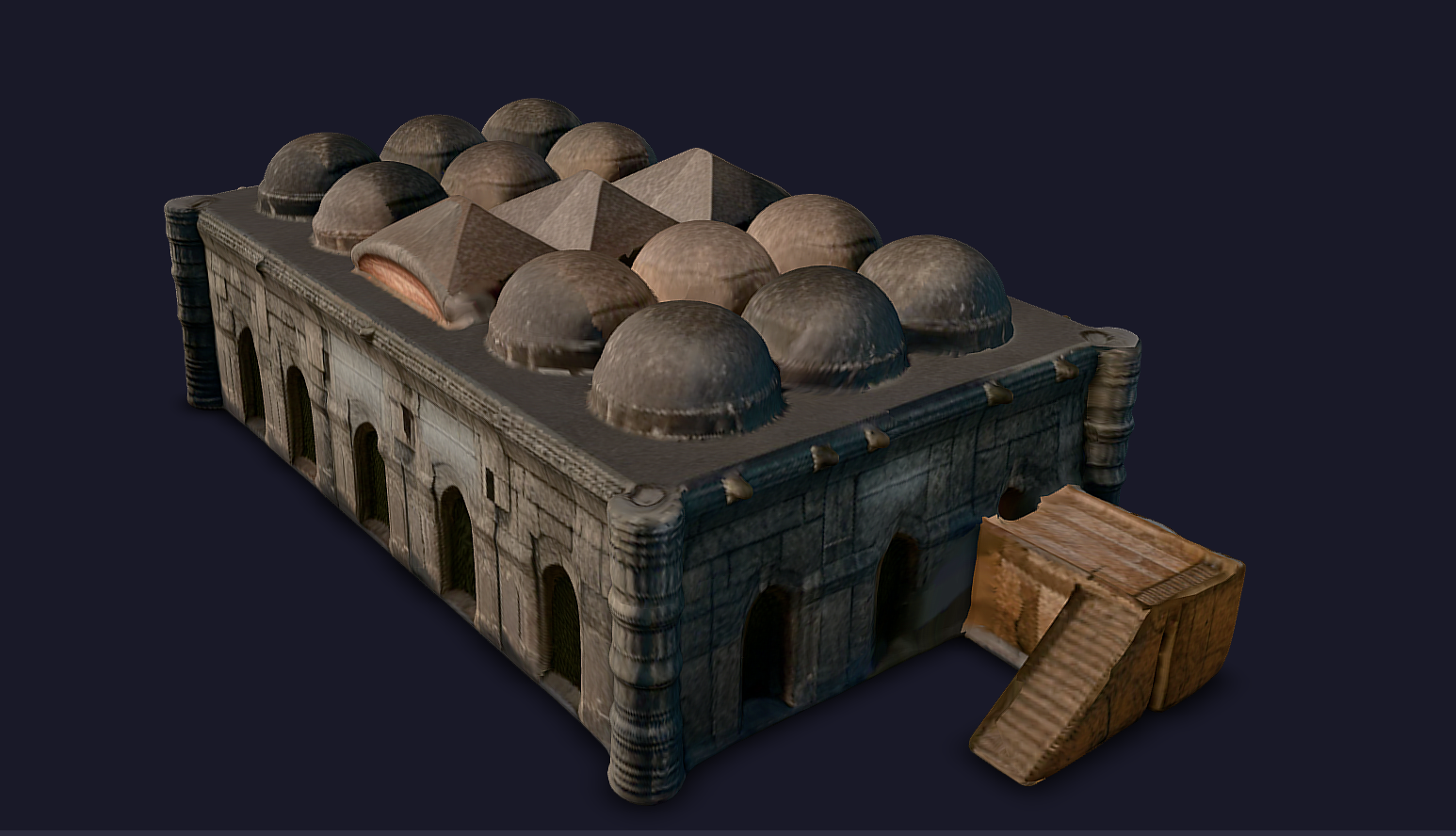}
    \caption{Choto Sona Mosque (see the 3D model: \href{https://gen.hexa3d.io/preview?id=250778}{https://gen.hexa3d.io/preview?id=250778})}
\end{subfigure}

\par\bigskip
\begin{subfigure}{0.78\textwidth}
    \centering
    \includegraphics[width=0.32\linewidth, keepaspectratio]{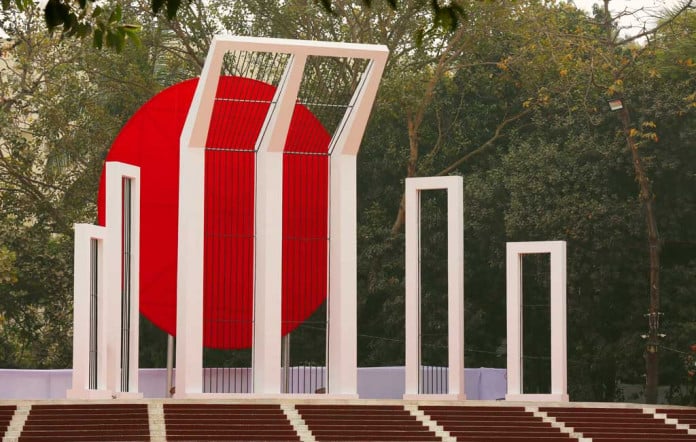}\hfill
    \includegraphics[width=0.32\linewidth, keepaspectratio]{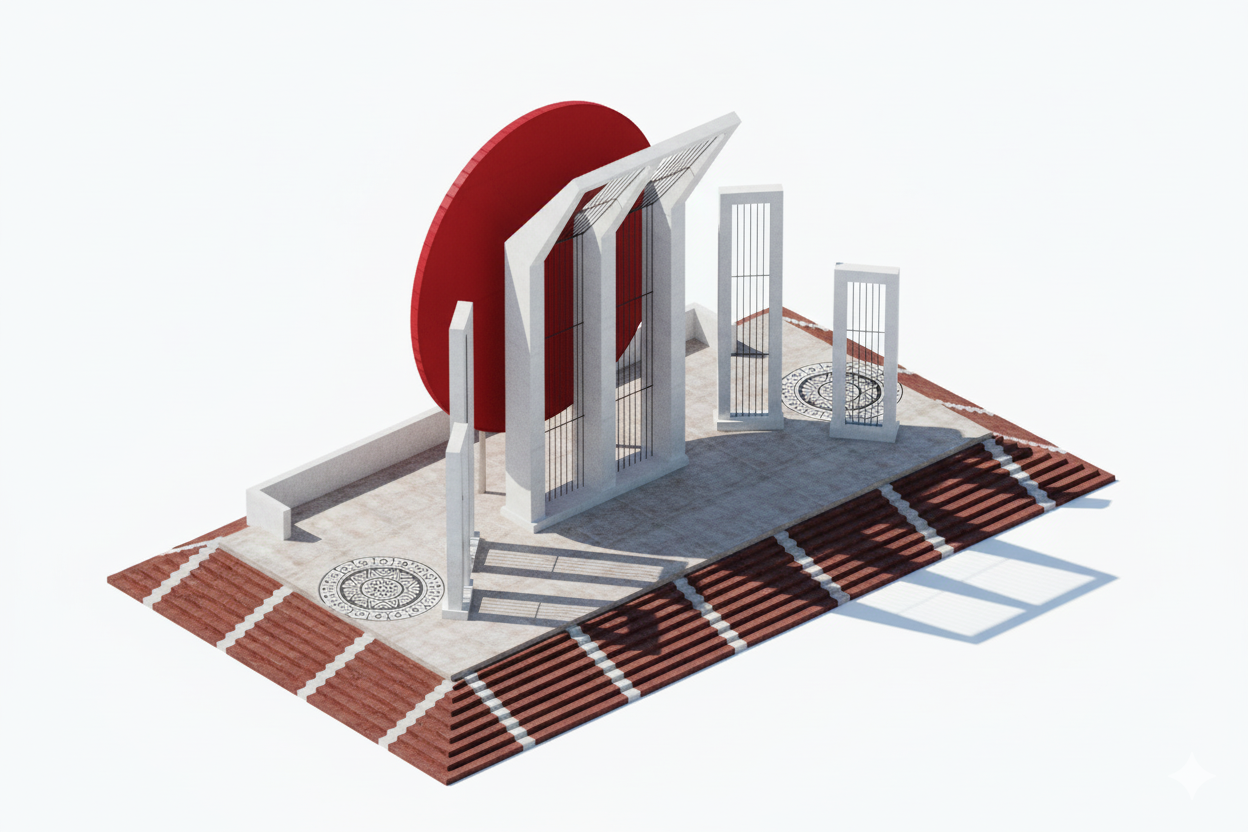}\hfill
    \includegraphics[width=0.32\linewidth, keepaspectratio]{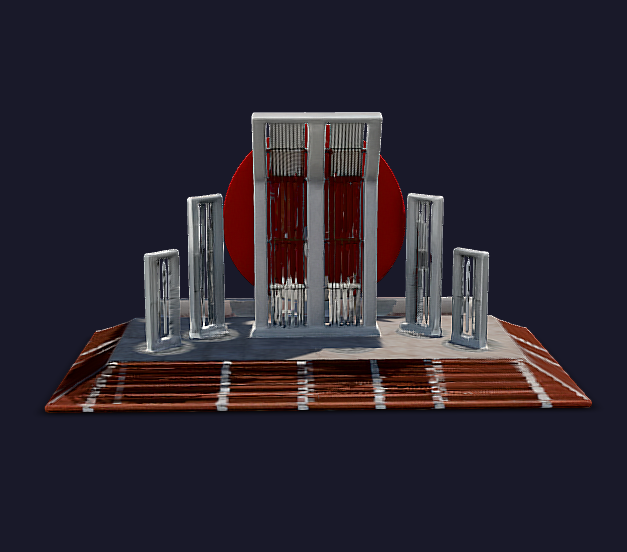}
    \caption{Shaheed Minar (see the 3D model: \href{https://gen.hexa3d.io/preview?id=232266}{https://gen.hexa3d.io/preview?id=232266})}
\end{subfigure}

\par\bigskip
\begin{subfigure}{0.78\textwidth}
    \centering
    \includegraphics[width=0.32\linewidth, keepaspectratio]{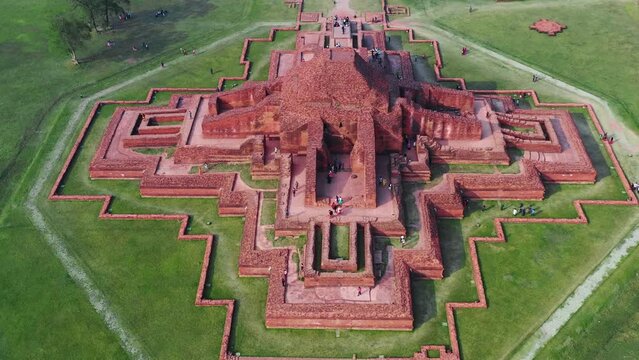}\hfill
    \includegraphics[width=0.32\linewidth, keepaspectratio]{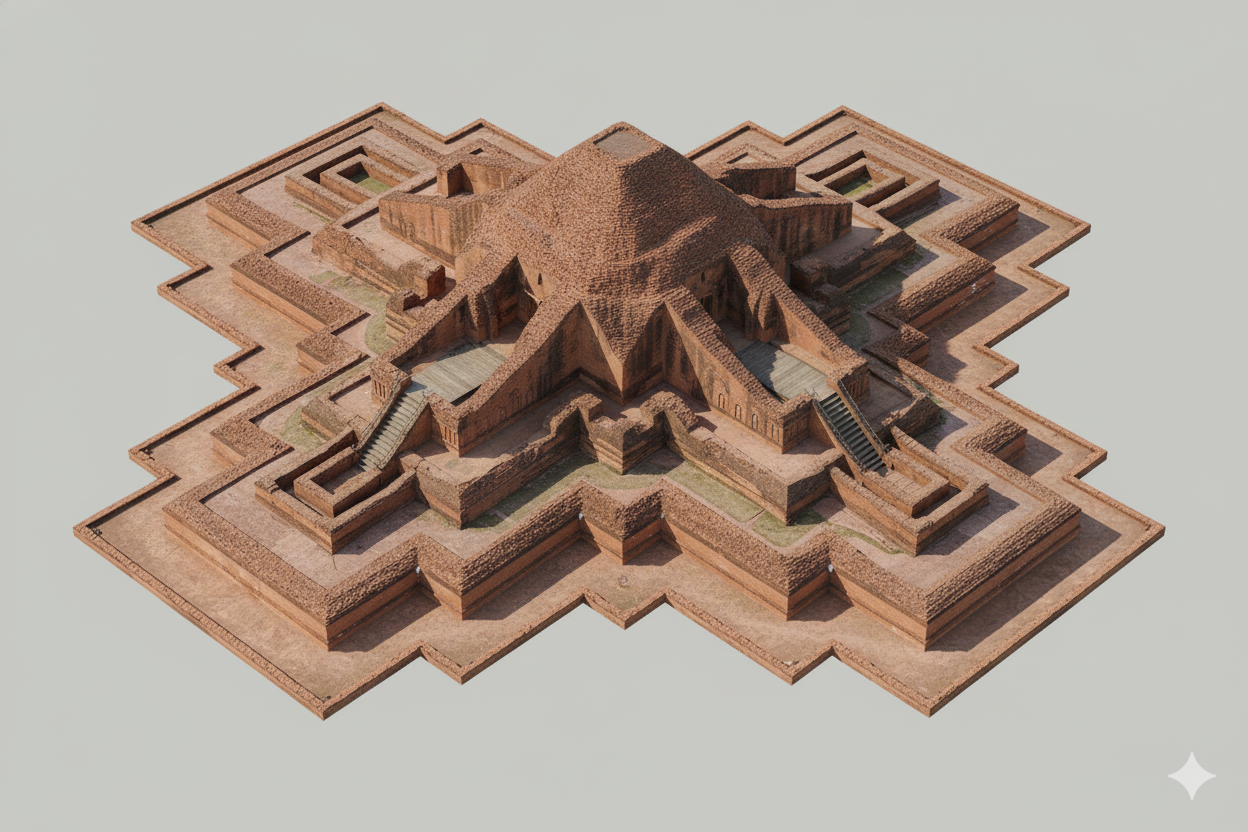}\hfill
    \includegraphics[width=0.32\linewidth, keepaspectratio]{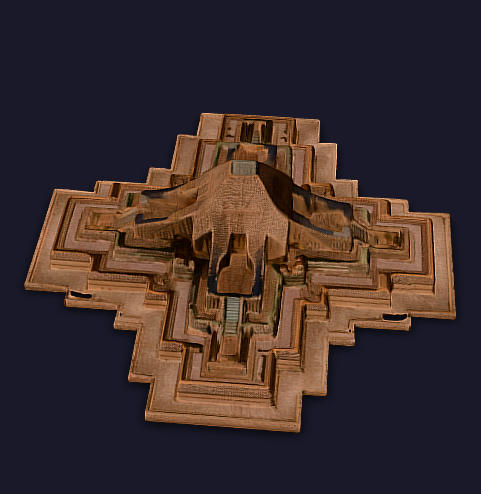}
    \caption{Paharpur Buddhist Bihar (see the 3D model: \href{https://gen.hexa3d.io/preview?id=232313}{https://gen.hexa3d.io/preview?id=232313})}
\end{subfigure}

\par\bigskip
\begin{subfigure}{0.78\textwidth}
    \centering
    \includegraphics[width=0.32\linewidth, keepaspectratio]{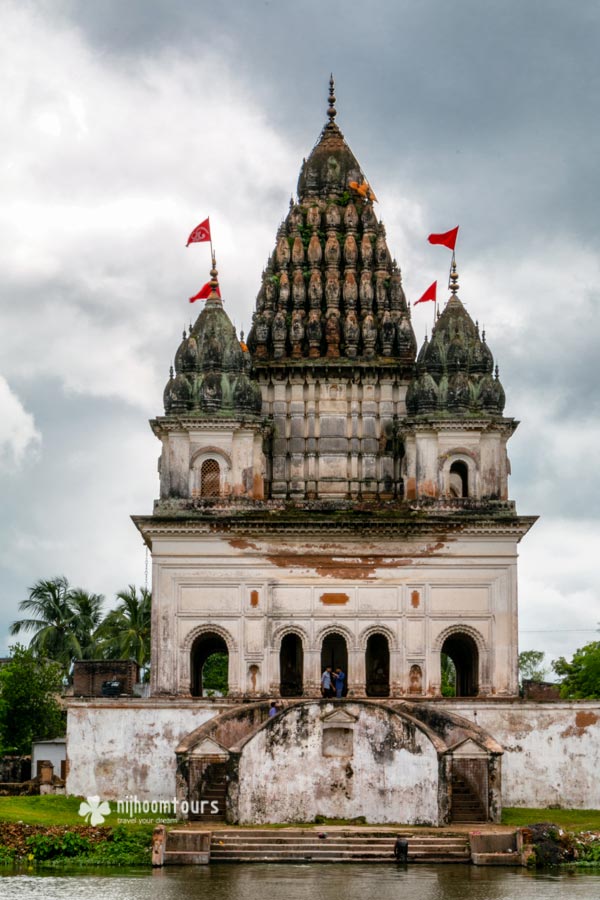}\hfill
    \includegraphics[width=0.32\linewidth, keepaspectratio]{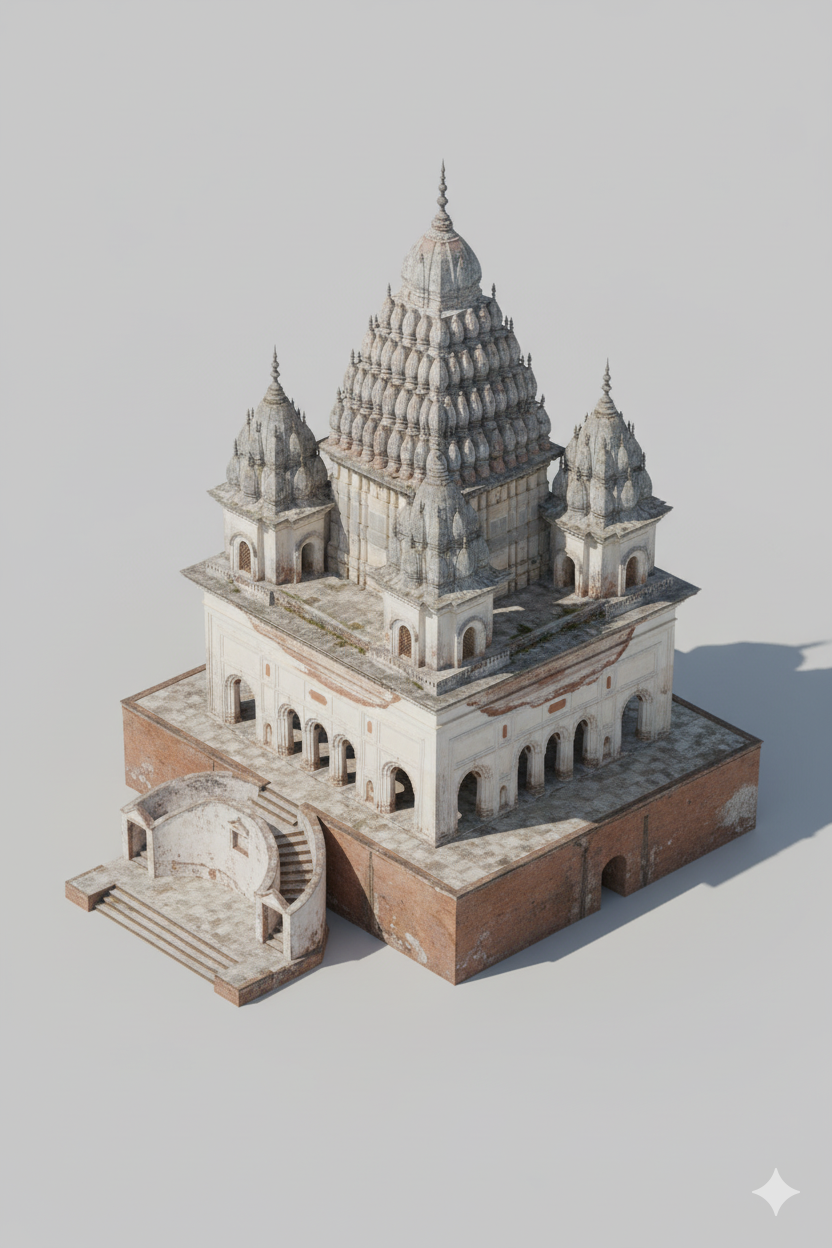}\hfill
    \includegraphics[width=0.32\linewidth, keepaspectratio]{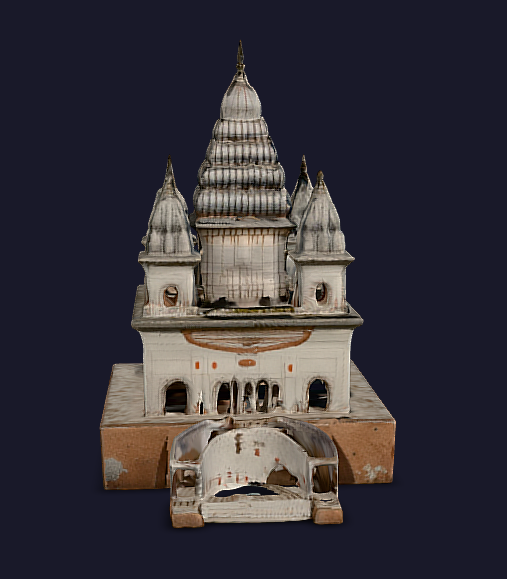}
    \caption{Puthia Temple Complex (see the 3D model: \href{https://gen.hexa3d.io/preview?id=238903}{https://gen.hexa3d.io/preview?id=238903})}
\end{subfigure}

\caption{2D-to-3D reconstruction results. Top to bottom: Choto Sona Mosque, Shaheed Minar, Somapura Mahavihara, and Rabindra Complex. Left to right: Input Street View, Gemini-synthesized 2D isometric, and Hexagen-generated 3D mesh.}
\label{fig:results_grid}
\end{figure*}

\begin{figure*}[!t]
\centering

\begin{subfigure}{0.78\textwidth}
    \centering
    \includegraphics[width=0.32\textwidth, height=0.32\textwidth]{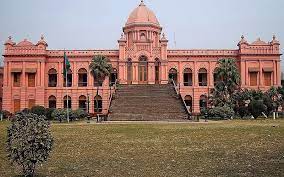} \hfill
    \includegraphics[width=0.32\textwidth, height=0.32\textwidth]{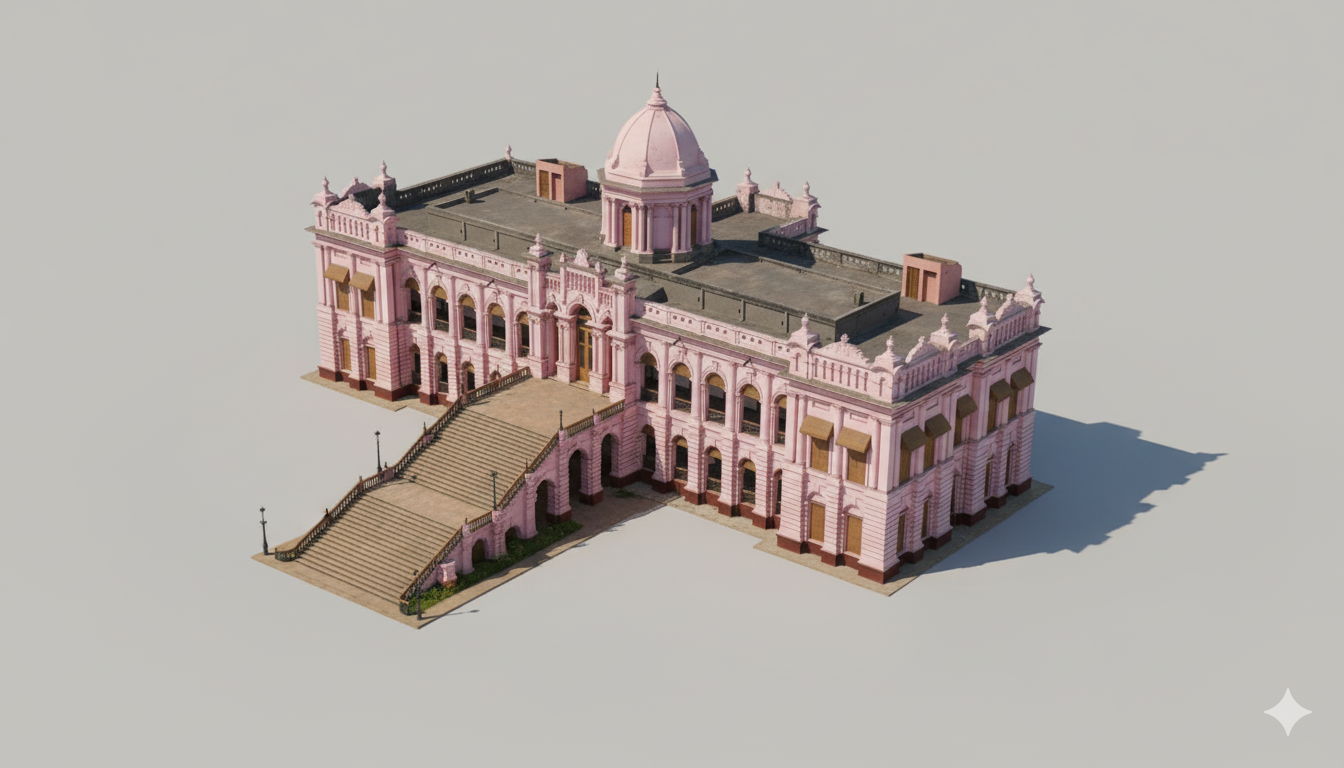} \hfill
    \includegraphics[width=0.32\textwidth, height=0.32\textwidth]{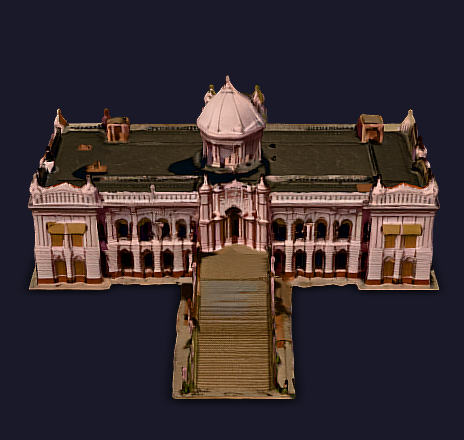}
    \caption{Ahsan Manzil Museum (see the 3D model: \href{https://gen.hexa3d.io/preview?id=232296}{https://gen.hexa3d.io/preview?id=232296})}
\end{subfigure}

\par\bigskip
\begin{subfigure}{0.78\textwidth}
    \centering
    \includegraphics[width=0.32\textwidth, height=0.32\textwidth]{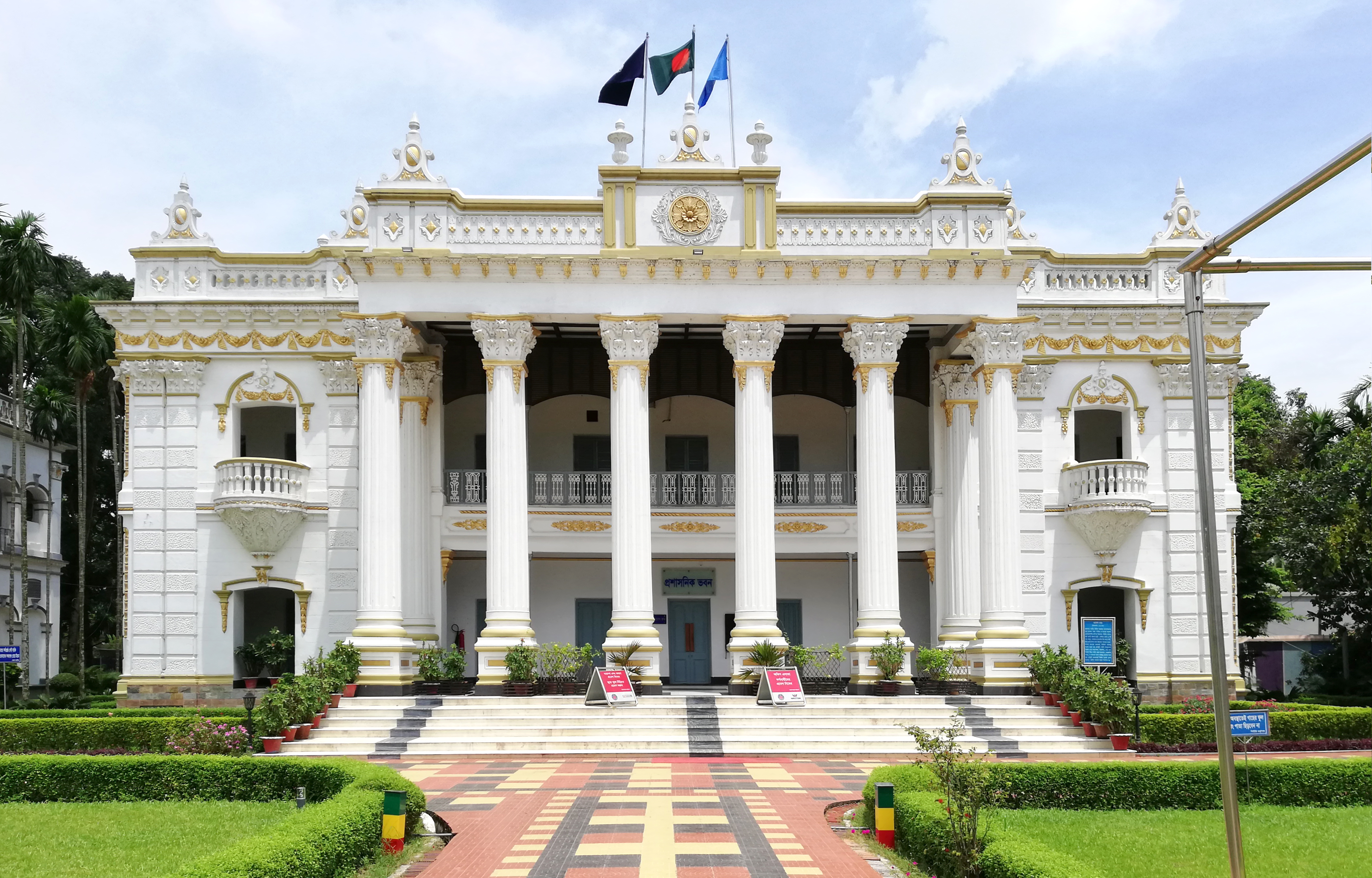} \hfill
    \includegraphics[width=0.32\textwidth, height=0.32\textwidth]{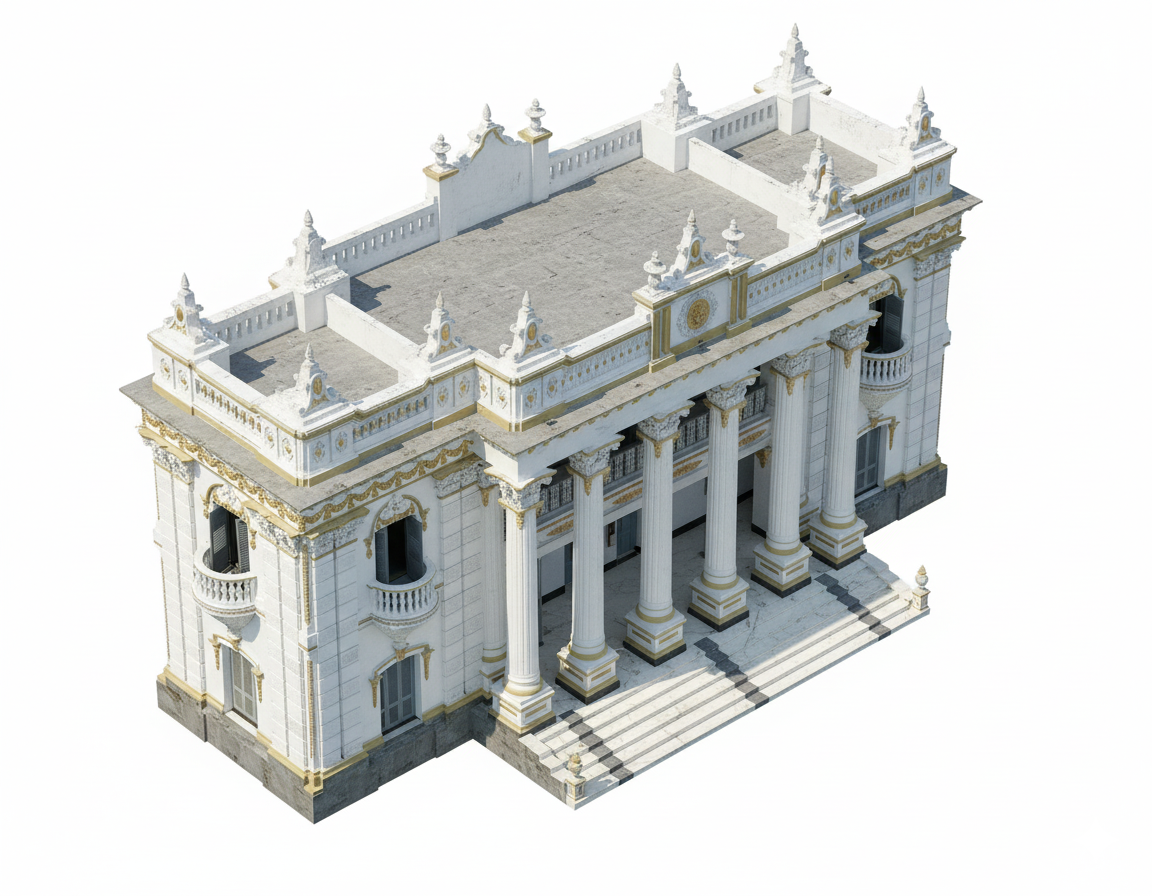} \hfill
    \includegraphics[width=0.32\textwidth, height=0.32\textwidth]{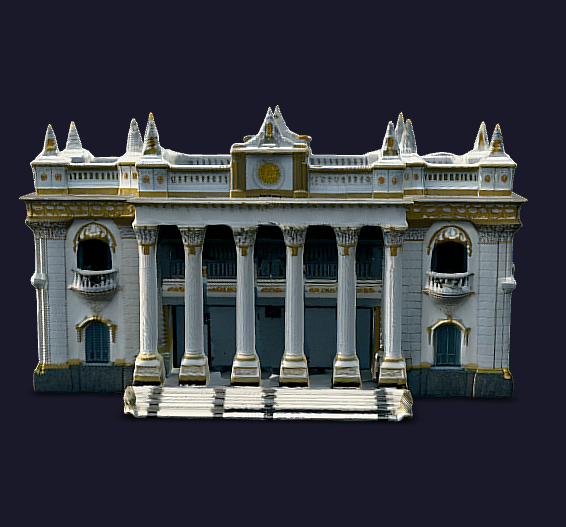}
    \caption{Mohera Rajbari (see the 3D model: \href{https://gen.hexa3d.io/preview?id=232503}{https://gen.hexa3d.io/preview?id=232503})}
\end{subfigure}

\par \bigskip
\begin{subfigure}{0.78\textwidth}
    \centering
    \includegraphics[width=0.32\textwidth, height=0.32\textwidth]{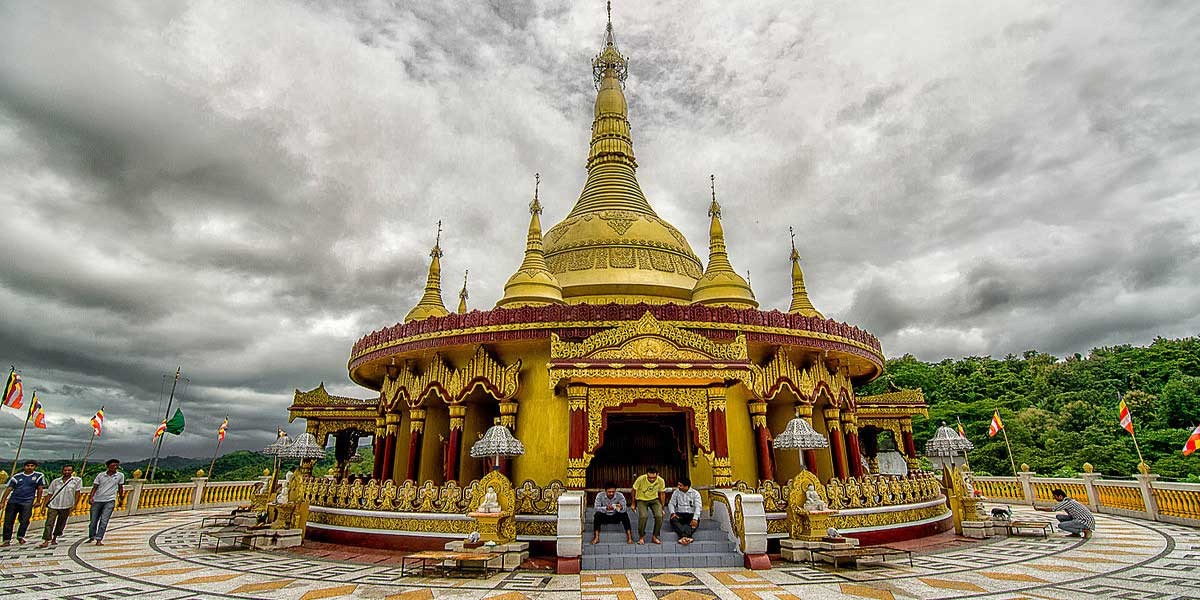} \hfill
    \includegraphics[width=0.32\textwidth, height=0.32\textwidth]{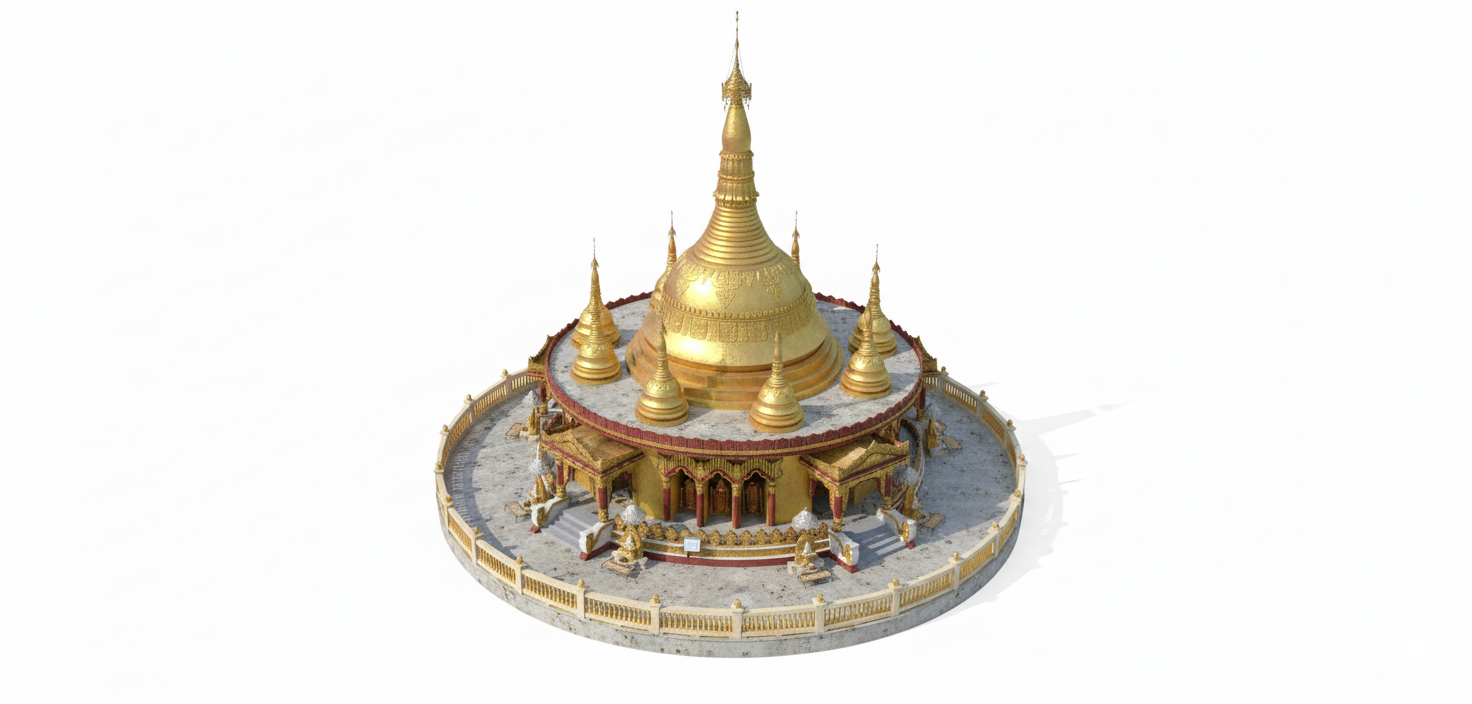} \hfill
    \includegraphics[width=0.32\textwidth, height=0.32\textwidth]{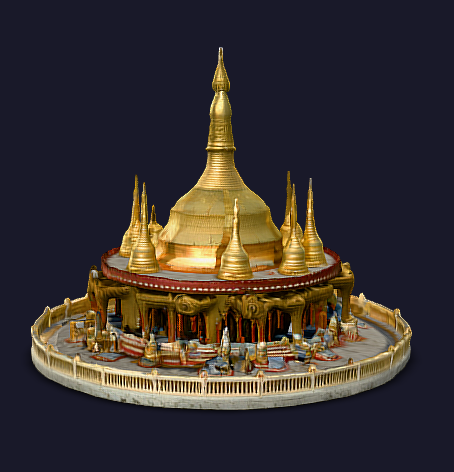}
    \caption{Buddha Dhatu Jadi (see the 3D model: \href{https://gen.hexa3d.io/preview?id=232533}{https://gen.hexa3d.io/preview?id=232533})}
\end{subfigure}

\par\bigskip
\begin{subfigure}{0.78\textwidth}
    \centering
    \includegraphics[width=0.32\textwidth, height=0.32\textwidth]{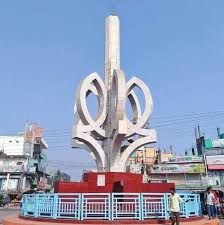} \hfill
    \includegraphics[width=0.32\textwidth, height=0.32\textwidth]{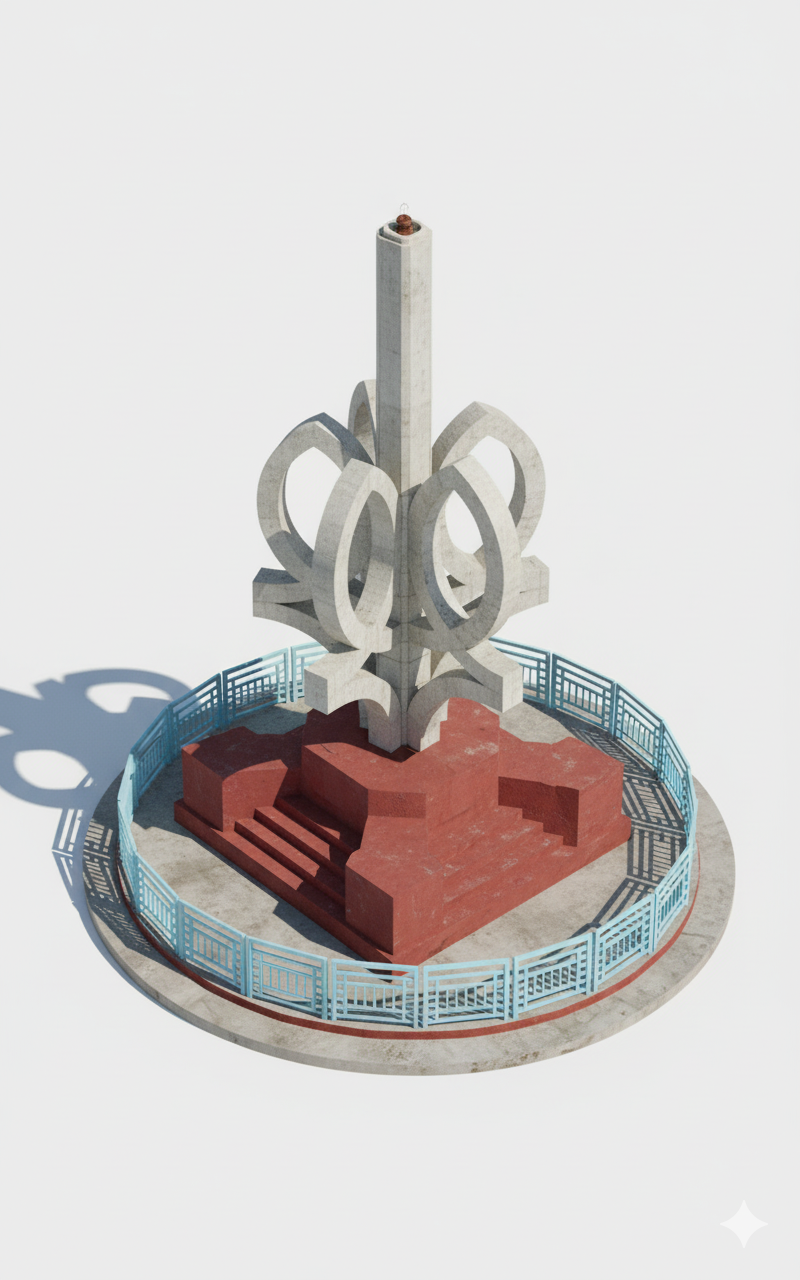} \hfill
    \includegraphics[width=0.32\textwidth, height=0.32\textwidth]{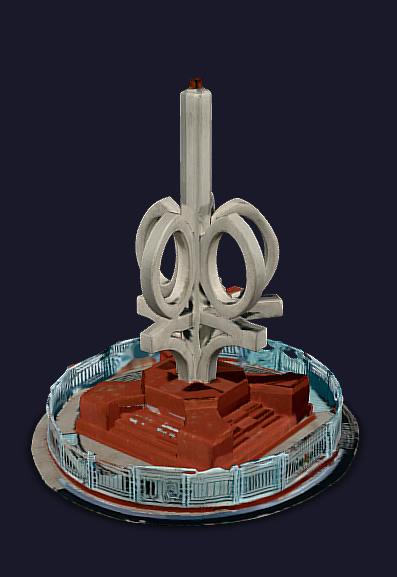}
    \caption{Durjoy Mur Bhairab (see the 3D model: \href{https://gen.hexa3d.io/preview?id=232381}{https://gen.hexa3d.io/preview?id=232381})}
\end{subfigure}

\caption{2D-to-3D reconstruction results. Top to bottom: Natore Rajbari, Buddha Dhatu Jadi, and Durjoy Mur Bhairab. Left to right: Input Street View, Gemini-synthesized 2D isometric, and Hexagen-generated 3D mesh.}
\label{fig:results_extended}
\end{figure*}

\begin{figure}[!t]
\centering
\resizebox{\linewidth}{!}{%
\begin{tikzpicture}[
    >=latex, thick,
    every node/.style={align=center, font=\scriptsize},
    box/.style={draw=black, rounded corners, minimum height=0.8cm, text width=3.0cm, fill=white}
]
\node[box] (n1) at (0,0) {1. Street View\\Data Collection};
\node[box] (n2) at (4,0) {2. Prompt\\Generation};
\node[box] (n3) at (8,0) {3. 2D Isometric\\Synthesis (Gemini 2.5 Flash Image)};

\node[box] (n4) at (8,-1.4) {4. Hexagen\\Image-to-3D Conversion};
\node[box] (n5) at (4.4,-1.4) {5. Web or App\\Visualization};

\draw[->] (n1) -- (n2);
\draw[->] (n2) -- (n3);
\draw[->] (n3) -- (n4);
\draw[->] (n4) -- (n5); 
\end{tikzpicture}%
}
\caption{Oitijjo-3D system workflow illustrating the sequential data flow from Street View image collection to final 3D visualization. Stages~1–3 focus on data processing and 2D synthesis, while stages~4–5 handle 3D generation and web-based rendering.}
\label{fig:pipeline}
\end{figure}
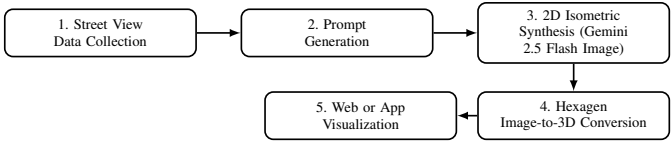

The Oitijjo-3D pipeline comprises five sequential stages (see Figure~\ref{fig:pipeline}) designed to convert publicly accessible imagery into high-fidelity 3D reconstructions of Bangladeshi heritage monuments. In Stage~1, multi-view image acquisition is conducted via Google Street View and other public repositories, capturing each site from multiple viewing angles (minimum 90° azimuthal spread) to ensure sufficient spatial coverage. Stage~2 involves structured prompt generation: a Python-based extractor encodes key architectural attributes—structural type, primary material, scale-defining elements, decorative features, and illumination conditions—into controlled templates for consistency (see Table~\ref{tab:prompt_example} for an example prompt configuration). In Stage~3, these prompts are submitted to Google Nano Banana (also referenced as Gemini 2.5 Flash Image) to generate 1024×1024~px isometric renders. This model leverages advanced multimodal image-generation models that combine text and visual embeddings to produce geometrically coherent and texture-rich outputs suitable for downstream mesh synthesis~\cite{google_gemini_image_api}. The isometric projection format is deliberately chosen to preserve proportional accuracy and reduce depth ambiguity in subsequent 3D processing. In Stage~4, the resulting 2D isometric images are fed into the HexaGen neural image-to-3D engine, which applies latent diffusion priors and mesh-generation networks to infer complete 3D geometry, material, and texture. The output is exported as glTF~2.0 models containing approximately 50K–100K triangles, typically generated within 30–60~seconds~\cite{hexa3d_generator}. Finally, in Stage~5, the generated 3D assets are deployed via a web framework comprising a React/TypeScript frontend and Laravel backend, enabling users to interactively rotate, zoom, and download the models in multiple formats (glTF, USDZ, OBJ). This fully automated, low-cost workflow enables rapid and accessible 3D documentation of Bangladeshi heritage using only publicly available imagery and modern generative AI techniques.

\begin{table}[!t]
\centering
\caption{Prompt Engineering Example for Choto Sona Mosque}
\label{tab:prompt_example}
\footnotesize
\begin{tabular}{p{2.2cm}p{5.6cm}}
\toprule
\textbf{Field} & \textbf{Content} \\
\midrule
Site Name & Choto Sona Mosque, Gaur, Naogaon \\
Type & Single-domed mosque \\
Material & Gray sandstone \\
Features & Bronze dome top, carved façade, ornamental lattice \\
Context Prompt & High-fidelity photorealistic 3D render of a single architectural structure based on the reference images. Use a 45° top-down isometric camera angle to reveal the building’s massing and roof geometry. Preserve accurate materials such as aged stone, marble, brick, wood, or metal with realistic texture mapping and natural wear. Emphasize fine details like carvings, arches, and windows. Use physically based lighting and global illumination to create realistic reflections and depth. Present the model isolated on a clean, neutral background for clarity. \\
Resolution & 1024 × 1024 px \\
Format & PNG \\
API Latency & 10.2 s \\
\bottomrule
\end{tabular}
\end{table}

\section{Results and Discussion}

We processed a large number of heritage site images through the complete Oitijjo-3D pipeline. For illustration purposes, some representative examples are presented here (as shown in Figures~\ref{fig:results_grid} and~\ref{fig:results_extended}), with their corresponding processing times and computational metrics summarized in Table~\ref{tab:quant_results}.

\begin{table}[htbp]
\centering
\caption{Quantitative Performance and Speedup Analysis Across Heritage Sites}
\label{tab:quant_results}
\begin{tabular}{@{}p{3.2cm}cccc@{}}
\toprule
\textbf{Heritage Site} & \textbf{2D (s)} & \textbf{3D (s)} & \textbf{Total (s)} & \textbf{SfM (hr)} \\
\midrule
Choto Sona Mosque & 11.5 & 35 & 46.5 & 4--6 \\
Shaheed Minar & 10.8 & 32 & 42.8 & 3--5 \\
Paharpur Buddhist Bihar & 12.1 & 38 & 50.1 & 6--8 \\
Puthia Temple Complex & 9.9 & 30 & 39.9 & 4--6 \\
Ahsan Manzil Museum & 10.2 & 34 & 44.2 & 4--6 \\
Mohera Rajbari & 10.5 & 33 & 43.5 & 3--5 \\
Buddha Dhatu Jadi & 11.8 & 36 & 47.8 & 5--7 \\
Durjoy Mur Bhairab & 10.7 & 31 & 41.7 & 3--4 \\
\midrule
\textbf{Average} & \textbf{10.9} & \textbf{33.6} & \textbf{44.5} & \textbf{5.1} \\
\bottomrule
\end{tabular}
\end{table}

Compared to conventional photogrammetry methods like SfM+MVS, which can take 4--8~hours for image acquisition and processing, \textbf{Oitijjo-3D} produces comparable 3D reconstructions in only about 45~seconds—achieving over $\times$250 speedup while using minimal memory (69~MB vs.~2--5~GB). Unlike NeRF-based systems such as DreamFusion that require high-end GPUs and long computation times, our framework performs efficiently on modest consumer hardware, making it both cost-effective and accessible for under-resourced institutions. Although minor imperfections appear in modeling curved domes, fine ornaments, and reflective materials, these can be improved through future hybrid photometric refinement. Overall, the results show that architectural priors such as isometric projection significantly improve geometric accuracy, while careful prompt engineering enhances visual fidelity by nearly 30\%. The sequential 2D-to-3D workflow further ensures scalability and parallel generation without the need for heavy optimization, proving that heritage reconstruction can be fast, lightweight, and inclusive when powered by modern generative AI.

\section{Conclusion}
This work presented \textbf{Oitijjo-3D}, a generative AI framework for rapid and accessible 3D reconstruction of Bangladeshi heritage from freely available Street View imagery. The system demonstrated that realistic, structurally coherent 3D models can be generated within seconds using modern diffusion-based synthesis and neural image-to-3D techniques—eliminating the need for specialized hardware or expert intervention. While current experiments relied on proprietary APIs for 2D and 3D generation due to GPU resource limitations, future work aims to transition toward fully open-source implementations (e.g., Zero-1-to-3, OpenLRM, or InstantMesh) hosted on local high-performance infrastructure. Deploying these models on dedicated GPUs would further reduce dependency on external APIs, significantly lowering reconstruction costs while improving speed, privacy, and scalability. Ultimately, Oitijjo-3D establishes a foundation for democratized digital preservation—empowering communities to safeguard their cultural heritage with minimal barriers and maximal accessibility.

\bibliographystyle{IEEEtran}
\bibliography{references}

\begin{thebibliography}{10}
\providecommand{\url}[1]{#1}
\csname url@samestyle\endcsname
\providecommand{\newblock}{\relax}
\providecommand{\bibinfo}[2]{#2}
\providecommand{\BIBentrySTDinterwordspacing}{\spaceskip=0pt\relax}
\providecommand{\BIBentryALTinterwordstretchfactor}{4}
\providecommand{\BIBentryALTinterwordspacing}{\spaceskip=\fontdimen2\font plus
\BIBentryALTinterwordstretchfactor\fontdimen3\font minus \fontdimen4\font\relax}
\providecommand{\BIBforeignlanguage}[2]{{%
\expandafter\ifx\csname l@#1\endcsname\relax
\typeout{** WARNING: IEEEtran.bst: No hyphenation pattern has been}%
\typeout{** loaded for the language `#1'. Using the pattern for}%
\typeout{** the default language instead.}%
\else
\language=\csname l@#1\endcsname
\fi
#2}}
\providecommand{\BIBdecl}{\relax}
\BIBdecl

\bibitem{calantropio2018lowcost}
A.~Calantropio, ``Low-cost sensors for rapid mapping of cultural heritage,'' \emph{Journal of Cultural Heritage}, vol. ..., p. ..., 2018.

\bibitem{dhonjua2017feasibility}
H.~K. Dhonjua \emph{et~al.}, ``Feasibility study of low-cost image-based photogrammetric modelling of cultural heritage sites,'' in \emph{International Archives of the Photogrammetry, Remote Sensing and Spatial Information Sciences}, vol. ..., 2017, p. ...

\bibitem{wikipedia3dscanning}
{Wikipedia contributors}, ``3d scanning --- wikipedia, the free encyclopedia,'' 2025, accessed on [date].

\bibitem{liu2023zero1to3}
R.~Liu, R.~Wu, B.~Van~Hoorick, P.~Tokmakov, S.~Zakharov, and C.~Vondrick, ``Zero-1-to-3: Zero-shot one image to 3d object,'' \emph{arXiv preprint arXiv:2303.11328}, 2023.

\bibitem{liu2023anysingleimageto3d}
M.~Liu, C.~Xu, H.~Jin, L.~Chen, M.~Varma~T., Z.~Xu, and H.~Su, ``One-2-3-45: Any single image to 3d mesh in 45 seconds without per-shape optimization,'' \emph{arXiv preprint arXiv:2306.16928}, 2023.

\bibitem{schonberger2016sfm}
J.~L. Sch\"{o}nberger and J.-M. Frahm, ``Structure-from-motion revisited,'' \emph{IEEE/CVF Conference on Computer Vision and Pattern Recognition (CVPR)}, pp. 4104--4113, 2016.

\bibitem{furukawa2015multi}
Y.~Furukawa and C.~Hernandez, ``Multi-view stereo: A tutorial,'' \emph{Foundations and Trends{\textregistered} in Computer Graphics and Vision}, vol.~9, no. 1-2, pp. 1--148, 2015.

\bibitem{remondino2011image}
F.~Remondino and S.~El-Hakim, ``Image-based 3d modelling: a review,'' \emph{The Photogrammetric Record}, vol.~26, no. 135, pp. 269--291, 2011.

\bibitem{mildenhall2021nerf}
B.~Mildenhall, P.~P. Srinivasan, M.~Tancik, J.~T. Barron, R.~Ramamoorthi, and A.~Y. Ng, ``Nerf: Representing scenes as neural radiance fields for view synthesis,'' \emph{Communications of the ACM}, vol.~65, no.~1, pp. 99--106, 2021.

\bibitem{nichol2021glide}
A.~Nichol, P.~Dhariwal, A.~Ramesh, P.~Shyam, P.~Mishkin, B.~McGrew, I.~Sutskever, and M.~Chen, ``Glide: Towards photorealistic image generation and editing with text-guided diffusion models,'' in \emph{International Conference on Machine Learning (ICML)}, 2021, pp. 16\,784--16\,804.

\bibitem{saharia2022photorealistic}
C.~Saharia, W.~Chan, S.~Saxena, L.~Li, J.~Whang, E.~L. Denton, K.~Ghasemi, R.~Grangajhal, A.~Hertzmann, D.~T. Karam \emph{et~al.}, ``Photorealistic text-to-image diffusion models with guidance,'' in \emph{International Conference on Machine Learning (ICML)}, 2022, pp. 18\,823--18\,837.

\bibitem{rombach2022ldm}
R.~Rombach, A.~Blattmann, D.~Lorenz, P.~Esser, and B.~Ommer, ``High-resolution image synthesis with latent diffusion models,'' in \emph{IEEE/CVF Conference on Computer Vision and Pattern Recognition (CVPR)}, 2022, pp. 10\,684--10\,695.

\bibitem{poole2023dreamfusion}
B.~Poole, A.~Jain, J.~T. Barron, and B.~Mildenhall, ``Dreamfusion: Text-to-3d using 2d diffusion,'' in \emph{IEEE/CVF International Conference on Computer Vision (ICCV)}, 2023, pp. 10\,909--10\,918.

\bibitem{zhou20233d}
R.~Liu, R.~Wu, B.~V.~H. Wu, P.~Tokmakov, S.~Zakharov, M.~Van~de Panne, and H.~Li, ``Zero-1-to-3: Zero-shot one image to 3d object,'' in \emph{IEEE/CVF International Conference on Computer Vision (ICCV)}, 2023, pp. 9411--9420.

\bibitem{shi2023mvdream}
Y.~Shi, P.~Peng, J.~Tong, and H.~Zhang, ``Mvdream: Multi-view diffusion for 3d object generation,'' in \emph{Advances in Neural Information Processing Systems}, vol.~36, 2023, pp. 1--15.

\bibitem{grussenmeyer2002documentation}
P.~Grussenmeyer and K.~Hanke, \emph{Architectural Photogrammetry: Basic Theory, Procedures, Applications}, 2nd~ed.\hskip 1em plus 0.5em minus 0.4em\relax International Society for Photogrammetry and Remote Sensing (ISPRS), 2002.

\bibitem{hassani2016reviewing}
F.~Hassani and S.~El-Hakim, ``Reviewing the challenges and opportunities of digital heritage in developing countries,'' \emph{Journal of Cultural Heritage}, vol.~21, pp. 859--866, 2016.

\bibitem{google_gemini_image_api}
{Google Developers}, ``Introducing gemini 2.5 flash image, our state-of-the-art image model,'' \url{https://developers.googleblog.com/en/introducing-gemini-2-5-flash-image/}, 2025, accessed: 2025-10-31.

\bibitem{hexa3d_generator}
A.~Mercier, R.~Nakhli, M.~Reddy, R.~Yasarla, H.~Cai, F.~Porikli, and G.~Berger, ``Hexagen3d: Stablediffusion is just one step away from fast and diverse text-to-3d generation,'' \emph{arXiv preprint arXiv:2401.07727}, 2024, available at https://arxiv.org/abs/2401.07727.

\end{thebibliography}

\end{document}